\documentclass[sigconf]{acmart}
\usepackage{subfigure}
\usepackage{graphicx}
\usepackage{enumitem}
\AtBeginDocument{%
  }

\copyrightyear{2023} 
\acmYear{2023} 
\setcopyright{acmlicensed}\acmConference[SIGIR '23]{Proceedings of the 46th International ACM SIGIR Conference on Research and Development in Information Retrieval}{July 23--27, 2023}{Taipei, Taiwan}

\begin{document}

\title{Dual Semantic Knowledge Composed Multimodal Dialog Systems}


\author{Xiaolin Chen}
\email{cxlicd@gmail.com}
\affiliation{%
  \institution{School of Software, Joint SDU-NTU Centre for Artificial Intelligence Research, Shandong University}
  \country{China}
}
\author{Xuemeng Song$^*$}
\email{sxmustc@gmail.com}
\affiliation{%
  \institution{School of Computer Science and Technology, Shandong University}
  \country{China}
}
\author{Yinwei Wei}
\email{weiyinwei@hotmail.com}
\affiliation{%
  \institution{School of Computing, National University of Singapore}
  \country{Singapore}
}
\author{Liqiang Nie$^*$}
\email{nieliqiang@gmail.com}
\affiliation{%
  \institution{School of Computer Science and Technology, Harbin Institute of Technology (Shenzhen)}
  \country{China}
}
\author{Tat-Seng Chua}
\email{dcscts@nus.edu.sg}
\orcid{0000-0003-4638-0603}
\affiliation{%
  \institution{School of Computing, National University of Singapore}
  \country{Singapore}
}
\thanks{$^*$Corresponding authors: Xuemeng Song and Liqiang Nie.}

\begin{abstract}
{\color{black}
Textual response generation  is an essential task for multimodal \mbox{task-oriented} dialog systems.
Although existing studies have achieved  fruitful progress, they still suffer from two critical limitations: \mbox{1) \emph{focusing}} \emph{on the attribute knowledge but ignoring the relation knowledge that can reveal the correlations between different entities and hence promote the response generation}, and 2) \emph{only conducting the cross-entropy loss based output-level supervision but lacking the representation-level regularization}. 
To address these limitations, we devise a novel multimodal \mbox{task-oriented} dialog system (named \mbox{MDS-S$^2$}).
Specifically, \mbox{MDS-S$^2$} first simultaneously acquires the context related attribute  and relation knowledge from the knowledge base, whereby the non-intuitive relation knowledge is extracted by the $n$-hop graph walk. 
Thereafter, considering that the attribute knowledge and relation knowledge can benefit the responding to different levels of questions, 
we design a multi-level knowledge composition module in MDS-S$^2$ to obtain the latent composed response representation. 
Moreover,  we devise a set of latent query variables to distill the semantic information from the composed response representation and the ground truth response representation, respectively, and thus conduct the \mbox{representation-level} semantic regularization.
Extensive experiments on a public dataset have verified the superiority of our proposed MDS-S$^2$. 
We have released the  codes and parameters to facilitate the research community.}
\end{abstract}



\maketitle

\section{Introduction}
\begin{figure}[!t]
    \centering
    \includegraphics[scale=0.29]{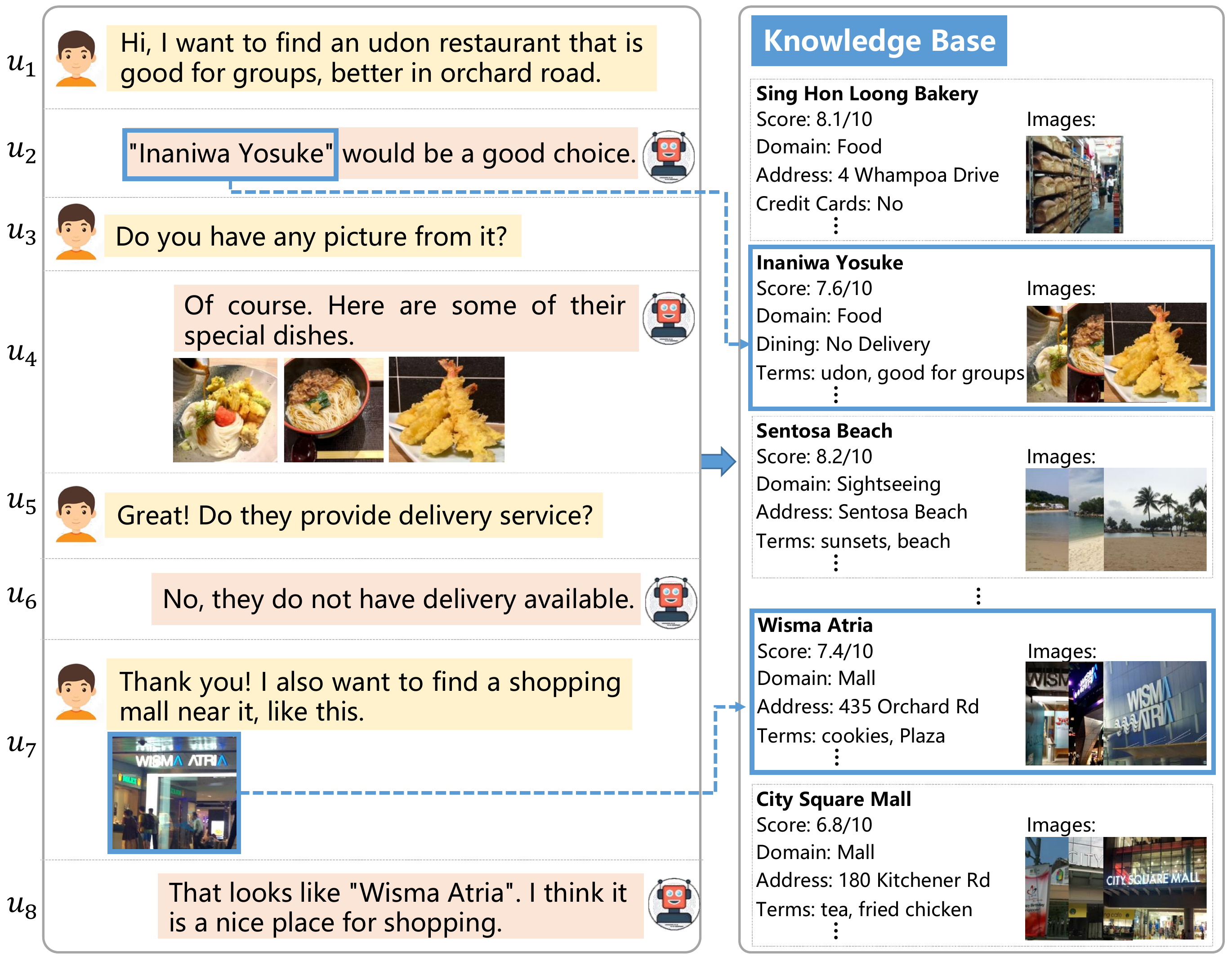}
    \vspace{-1em}
    \caption{{\color{black}Illustration of a multimodal dialog system between a user and an agent. ``u'' refers to the utterance.}}
    \vspace{-1.2em}
    \label{MMD_example}
\end{figure}

In recent years,  task-oriented dialog systems have penetrated into many aspects of our daily life,  
such as  restaurant  reserving and ticket booking. 
According to the report of Salesforce\footnote{https://startupbonsai.com/chatbot-statistics.}, roughly $68\%$ of customers tend to interact with the intelligent dialog agents for their quick responses rather than waiting for the human services.
Considering its value,  a surge of researches are dedicated to developing  task-oriented dialog systems. Early studies in this research line focus on the pure text-based dialog system~\cite{DBLP:conf/sigir/HeDYSHSL22,DBLP:conf/sigir/SiroAR22}, overlooking that both the user and the agent may need to express themselves with certain visual clues~(\emph{i.e.,} images).
For example, as shown in Figure~\ref{MMD_example}, the user needs to utilize the image to express his/her desired shopping mall in the utterance $u_7$, while the agent needs to use images to illustrate special dishes for the user in $u_4$.
Therefore, recent research attention has been swifted to the multimodal task-oriented dialog systems.

In fact, multimodal task-oriented dialog systems mainly contain two tasks~\cite{DBLP:journals/corr/abs-2207-07934}: the textual response generation and the image response selection.
Considering that the former is more challenging and its  performance is  still far from satisfactory, many researchers focus on this task for multimodal task-oriented dialog systems~\cite{DBLP:conf/mm/LiaoM0HC18, ma-etal-2022-unitranser}.
Despite the favorable  performance obtained by existing efforts~\cite{DBLP:conf/mm/LiaoM0HC18, ma-etal-2022-unitranser,DBLP:conf/aaai/SahaKS18,DBLP:journals/corr/abs-2207-07934,DBLP:journals/tip/NieJWWT21,DBLP:conf/sigir/CuiWSHXN19,DBLP:conf/mm/ZhangLGLWN21}, 
they mainly suffer from two critical limitations.
1) \textbf{Ignoring the relation knowledge.} In the context of  multimodal task-oriented dialog systems, there is always a knowledge base containing   
abundant attribute-value pairs as well as images of a large number of entities.
Previous studies focus on exploiting the attribute knowledge of entities, but neglecting the relation knowledge  residing in the knowledge base, which can capture the relations between entities, and benefit the response reasoning and generation. For example, as shown in Figure~\ref{MMD_example}, the agent can generate the appropriate response~(\emph{i.e.,} $u_8$) only conditioned on the relation knowledge of Inaniwa Yosuke $\xrightarrow{near}$  Wisma Atria $\xrightarrow{domain}$  mall. 
{\color{black}2) \textbf{Lacking the \mbox{representation-level} regularization.} Previous studies only adopt the token-level \mbox{cross-entropy} loss to regulate the generated response to be similar to the ground truth response.
This may be insufficient for the task whose input~(\emph{i.e.,} text and image) and output~(\emph{i.e.,} text) 
present apparent 
heterogeneity. 
{\color{black}In fact, 
they ignore the potential representation-level regularization between the context-knowledge composed response representation and the ground truth response representation, which can enhance the composed response representation learning and hence improve the response generation performance.} }

To address these two limitations, in this work, we aim to conduct the research of textual response generation  in multimodal task-oriented dialog systems by integrating the dual semantic  knowledge~(\emph{i.e.,} attribute and relation knowledge) and the representation-level regularization.
However, this is non-trivial
due to the following three challenges.
1) Different from the attribute knowledge,  the relation one is not straightforwardly provided by the knowledge base that only contains \mbox{attribute-value} pairs and images of entities. Hence, how to mine 
the related relation knowledge 
with respect to 
the given multimodal context is a crucial challenge.
{\color{black}\mbox{2) In} a sense, the intuitive attribute knowledge is 
 beneficial to  response generation of   simple questions~(\emph{e.g.,} ``Can you get their phone number for me?''), while the relation knowledge is helpful for responding relatively more complicated questions~(\emph{e.g.,} ``Can you help me look for a hotel nearby Singapore River?'').
Accordingly, how to effectively compose the multimodal context with the dual semantic knowledge and thus generate the proper response is another vital challenge.}
{\color{black}And 3) ideally, we expect that the representation-level regularization can project the \mbox{context-knowledge} composed response representation and the ground truth response representation into the same meaningful semantic space.
In this way, we can yield meaningful composed response representation that can  further enhance the response generation. }
Therefore, how to fulfil the meaningful \mbox{representation-level} semantic regularization is another challenge.

\begin{figure*}[!t]
    \centering
    \includegraphics[scale=0.44]{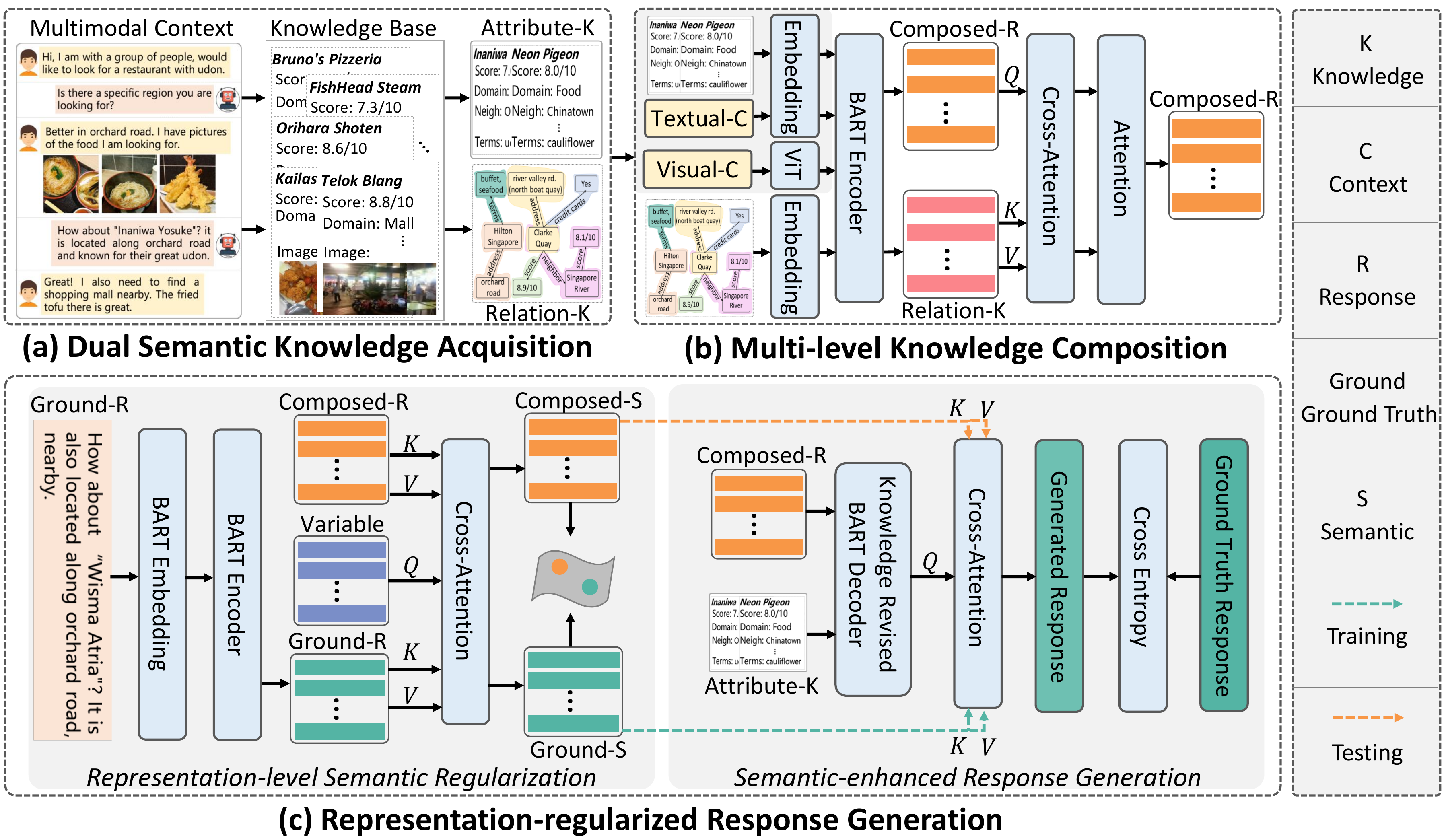}
    \caption{\color{black}{Illustration of the proposed model.}
}
    \vspace{-0.8em}
    \label{Model_figure}
\end{figure*}

 {\color{black}To address the aforementioned challenges, we devise a novel 
 dual semantic knowledge composed multimodal dialog system, 
 \mbox{MDS-S$^2$} for short, where the generative pretrained language model BART is adopted as the backbone.}
As demonstrated in Figure~\ref{Model_figure},
the proposed model consists of three pivotal components: 
\emph{dual semantic knowledge acquisition}, \emph{multi-level knowledge composition}, and \emph{\mbox{representation-regularized} response generation}.
To be specific, the first component aims to acquire the context related dual semantic knowledge: attribute knowledge and relation knowledge. In particular, the related relation knowledge is uncovered by the $n$-hop graph walk over the whole knowledge base. 
Thereafter, the second component is devised to compose the multimodal context and the acquired dual semantic knowledge to obtain the latent composed response representation.
Specifically,  {\color{black}considering that the attribute
knowledge and relation knowledge can facilitate the responding to questions with different complexity levels of user intentions,}
the attribute knowledge is first composed at the input token level,  and the relation knowledge is then adaptively composed at the intermediate representation level.
Subsequently, the last component
targets at enhancing the proper textual response generation with the additional \mbox{representation-level} regularization.   In particular, we design a set of \mbox{to-be-learned} latent query variables to project the composed response representation and the ground truth response representation into the same semantic space with the cross-attention mechanism.
Moreover, to fully utilize the \mbox{representation-level} regularization, we also design a \mbox{semantic-enhanced} response decoder.
Notably, the decoder can  adaptively incorporate  the regularized composed response semantic representation, apart from the  original multi-level knowledge composed response representation.
Extensive experiments on a public dataset have demonstrated the superiority of our proposed \mbox{MDS-S$^2$}.
Our main contributions can be summarized as follows:
\begin{itemize}[leftmargin=*]
    \item {\color{black}We propose a novel dual semantic knowledge composed multimodal dialog system. 
    To the best of our knowledge, we are among the first to exploit the relation knowledge residing in the knowledge base and integrate the \mbox{representation-level} semantic regularization for the textual response generation in multimodal task-oriented dialog systems.}
    \item We present the dual semantic knowledge acquisition component to select  the context related knowledge from both the attribute and relation perspectives. Moreover, we devise the multi-level knowledge composition component, which can compose the attribute  and  relation knowledge  at the input token level and the intermediate representation level, respectively.
    \item
    We devise a set of to-be-learned latent variables to conduct the representation-level semantic regularization, and the
\mbox{semantic-enhanced} response decoder to promote the textual response generation.
    As a byproduct, we release the codes and involved parameters to facilitate the research community\footnote{https://sigir2023.wixsite.com/anonymous7357.}.
\end{itemize}

\section{Related Work}
{\color{black}
Traditional task-oriented dialog systems~\cite{DBLP:conf/sigir/SongYBWWXWY21,DBLP:conf/sigir/LeiZSLMLYC22} resort to a pipeline structure and mainly contain four  functional components: natural language understanding, dialogue state tracking, policy learning, and natural language generation.
To be more specific, the natural language understanding component is used to classify  the user intention, and then the dialogue state tracking component aims to track the immediate state, based on which the policy learning component can predict the following action.
Thereafter,
the natural language generation component exhibits  the response through generation methods~\cite{DBLP:conf/sigir/HeDYSHSL22,DBLP:conf/sigir/LiLW21} %
or predefined templates.
Although pipeline methods have attained impressive results, they may 
suffer from   error propagation~\cite{DBLP:conf/acl/KanHLJRY18} 
on the sequential components.

With the flourishing development of deep neural networks, recent studies are  centered on exploring end-to-end \mbox{task-oriented} dialog systems.
Early end-to-end studies 
focus on  single-modal~(\emph{i.e.,} textual modality) \mbox{task-oriented} dialog systems.
Although these studies have made  tremendous strides, they neglect that 
both the user and agent may need to leverage  certain images to convey their needs or services.
Accordingly, Saha et al.~\cite{DBLP:conf/aaai/SahaKS18} investigated the multimodal  task-oriented dialog systems with two critical tasks: textual response generation and image response selection, and presented a multimodal hierarchical encoder-decoder model (MHRED).
In addition, they released a large-scale multimodal dialog dataset in the fashion domain, which considerably stimulates the  progress on multimodal task-oriented dialog systems.
{\color{black}Beyond this, several studies further probe the semantic relation in the multimodal dialog context and integrate the knowledge based on the framework of MHRED~\cite{DBLP:conf/acl/ChauhanFEB19,DBLP:conf/sigir/CuiWSHXN19,DBLP:conf/mm/LiaoM0HC18,DBLP:journals/tip/NieJWWT21,10.1145/3343031.3350923,DBLP:conf/mm/ZhangLGLWN21}.}
More recently, several studies draw on Transformer~\cite{DBLP:conf/nips/VaswaniSPUJGKP17} to propel the development of multimodal dialog systems~\cite{10.1145/3394171.3413679,DBLP:conf/acl/MaL0C22,DBLP:journals/corr/abs-2207-07934}.
Although these studies achieve
remarkable performance, they only utilize the attribute knowledge and overlook the representation-level regularization.
Beyond that, in this paper, we worked on investigating the 
dual semantic knowledge composition and representation-level semantic regularization to improve the response generation performance.}
\section{Model}


\subsection{Problem Formulation}
Suppose we have a set of $N$ training dialog pairs $\mathcal{D}=\{({\mathcal{C}_1},{\mathcal{R}_1}),({\mathcal{C}_2},{\mathcal{R}_2}),\cdots,({\mathcal{C}_N},{\mathcal{R}_N})\}$. 
Thereinto, each dialog pair consists of a multimodal dialog context $\mathcal{C}_i$ and a ground truth response $\mathcal{R}_i$.
In particular, each utterance in $\mathcal{C}_i$ may involve both textual and visual modalities, as the user/agent may utilize certain related images to promote the request/response expression. 
In light of this, each multimodal dialog context $\mathcal{C}_i$ can be further represented  by two modalities: 
the sequence of tokens $\mathcal{T}_i=[t^i_g]_{g=1}^{N_T^i}$  derived by concatenating all the textual utterances in the context 
and a set of images $\mathcal{V}_i=\{v_j^i\}_{j=1}^{N_V^i}$ involved in the context, where $t^i_g$ refers to the $g$-th token and $v_j^i$ represents the $j$-th image of $\mathcal{C}_i$. $N_T^i$ and $N_V^i$ are the total number of tokens and images, respectively. 
Notably, $N_V^i=0$~(\emph{i.e.,} $\mathcal{V}_i=\varnothing$), if there is no  image involved in $\mathcal{C}_i$.
The ground truth response $\mathcal{R}_i$ can be represented as  $\mathcal{R}_i=[r^i_n]^{N_R^i}_{n=1}$, where $r^i_n$ stands for the $n$-th token and $N_R^i$ is the number of tokens in the response.
In addition,
we have a knowledge base
including both the semantic and visual knowledge of  
$N_K$ entities $\mathcal{K}=\{e_p\}_{p=1}^{N_K}$ to assist in 
 response generation. 
To be specific, each entity $e_p$ is associated with a set of \mbox{attribute-value} pairs  $\mathcal{A}_p$~(\emph{e.g.,} \{<location: Orchard Road>, <domain: food>\}) and a set of images $\mathcal{I}_p$
that exhibit  
the visual information of the entity (\emph{e.g.,} showing the appearance of  Esplanade Park).

In a sense, we aim to devise a novel model $\mathcal{F}$ which can  generate the appropriate textual response based on the given multimodal  dialog context and the knowledge base as follows,
\begin{equation}
    \mathcal{F}(\mathcal{C}_i, \mathcal{K}|\boldsymbol{\Theta}_F)\rightarrow{\mathcal{R}_i},
     \label{eq1}
 \end{equation}
 where ${\boldsymbol{\Theta}_F}$ denotes the model parameters.

\subsection{Dual Semantic Knowledge Acquisition}
Since  knowledge plays a vital role in the response generation of \mbox{task-oriented} dialog systems, 
we first conduct the knowledge acquisition for the given multimodal context. 
Considering  the semantic knowledge is pivotal to capturing the user's intentions~\cite{DBLP:conf/mm/ZhangLGLWN21,wei2022causal,DBLP:conf/ijcai/YangZEL21}, we  focus on selecting two kinds of semantic knowledge: attribute knowledge and relation knowledge. 
Thereinto, the attribute knowledge, which is widely used, 
refers to the  \mbox{attribute-value} pairs of entities mentioned directly in the context.
In a sense, the attribute knowledge 
can be useful for 
responding the simple questions, like ``Can you get their phone number for me?''. 
In addition to the 
attribute knowledge, beyond previous methods, we also incorporate the relation knowledge contained in the knowledge base, 
which helps to uncover the correlation between entities and respond to  relatively more complicated questions, like ``Can you help me look for a hotel nearby Esplanade Park to stay at?''. 
{\color{black}To answer this complicated question, we need to first identify which entities are near the entity ``Esplanade Park'', and then recognize which nearby entity is a ``hotel'' with the attribute knowledge of nearby entities. }
Therefore, 
we devise the dual semantic  knowledge acquisition  component with two  modules: \emph{attribute knowledge acquisition} and \emph{relation knowledge acquisition}.

\subsubsection{Attribute Knowledge Acquisition.}
Due to the multimodal nature of the given dialog context, following the existing method~\cite{DBLP:journals/corr/abs-2207-07934}, we retrieve the related attribute knowledge  according to both the textual and visual context.

For the textual context, we directly judge which knowledge entity is mentioned to obtain the textual context related knowledge entities.
Namely, if the knowledge  entity $e_p$ appears in the textual context, we will select the set of  \mbox{attribute-value} pairs $\mathcal{A}_p$ of it as the relevant knowledge.
In this way, we can collect the related knowledge of textual context  $\mathcal{K}_t^A =\mathcal{A}_1^t \cup \mathcal{A}_2^t \cup \cdots\cup \mathcal{A}_{N_k^t}^t$, where $\mathcal{A}_m^t$ is the set of  \mbox{attribute-value} pairs  of the \mbox{$m$-th} mentioned knowledge entity  and $N_k^t$ is the total number of textual context related knowledge entities.

{\color{black}Pertaining to the visual context, 
we resort to mining its visual features to obtain the related entities in the knowledge base.
Specifically,
we first utilize \mbox{ViT-B/32}~\cite{DBLP:conf/iclr/DosovitskiyB0WZ21} to derive the visual features of images in   context $\mathcal{C}_i$ and that of  each entity in the knowledge base~$\mathcal{K}$.}
Notably, similar to the visual context, each entity in the knowledge base can be associated with multiple images. 
Thereafter,  for each image $v_j$ in the visual context $\mathcal{V}_i$, we calculate its visual similarity\footnote{Here, we use the cosine similarity.} with each image of each entity 
in the knowledge base
and regard the maximum image similarity as the  entity  similarity with the given image.
Notably, to guarantee the quality of the retrieved knowledge,
we only regard the  entities whose  similarity with the given context image is larger than the threshold $\epsilon$ as the related entities.
Ultimately, by merging the attribute-value pairs of all the visual context related entities, 
we obtain the visual context related attribute knowledge as  $\mathcal{K}_v^A=\mathcal{A}^v_1 \cup \mathcal{A}^v_2\cup \cdots \cup \mathcal{A}^v_{N_k^v}$, 
where $\mathcal{A}^v_n$ denotes the \mbox{attribute-value} pairs set 
of the $n$-th related knowledge entity  and ${N_k^v}$ is the total number of visual context related knowledge entities.

\subsubsection{Relation Knowledge Acquisition.}
To obtain the related relation knowledge, 
we first  cast the whole knowledge base $\mathcal{K}$ into a directed knowledge graph 
$\mathcal{G}_a=\{\mathcal{E}_a, \mathcal{R}_a\}$.  
$\mathcal{E}_a=\{e_q\}_{q=1}^{N_a^K}$ is the set of nodes, including two types of nodes~(\emph{i.e.,} head and tail nodes), where the head node refers to a knowledge entity, while the tail node denotes an attribute value.
$N_a^K$ is the total number of nodes.
 ${\mathcal{R}_a=\{r_z\}_{z=1}^{N_a^R}}$ is the  edge set, where each $r_z$ refers to an attribute type linking the head and tail nodes. $N_a^R$ is the number of attribute types in the knowledge base.
Intuitively, each triplet $(h, {r}, t)$, where $h, t\in\mathcal{E}_a$ and $r\in \mathcal{R}_a$,  
indicates that the attribute value of the knowledge entity $h$ regarding the attribute type $r$ is ${t}$.
For example, the triplet (\emph{Wisma Atria, location, Orchard Road}) indicates that the ``location'' of the entity ``Wisma Atria'' is at ``Orchard Road''.

Thereafter, for the given multimodal dialog context, 
we first identify the entities involved in the given context in  the same way as the attribute knowledge acquisition module.
Then, for each identified entity, we perform the $n$-hop graph walk over $\mathcal{G}_a$ to 
uncover the potential relations it involves.
In particular,  each hop walk  traverses a triplet~(\emph{i.e.,} $<$\hspace{0em}${e}, {r}, {\bar{e}}$\hspace{0em}$>$) and the  $n$-hop graph walk process will terminate when 
the last traversed node
of the current walk does not connect to any other nodes,
or the number of walks reaches the pre-defined maximum number~(\emph{i.e.,} $n$).
Finally, each $n$-hop graph walk yields a sequence of traversed triplets, which can compose a high-order relation of the initial entity node. 
Formally, we use a tuple to represent each relation, whose entries include the sequence of traversed nodes and edges. For example, the high-order relation $e_1\xrightarrow{r_1} e_2\xrightarrow{r_2} e_3$ can be represented as $[e_1, r_1, e_2, r_2, e_3]$. 
In this vein, we can derive all the context related relation  tuples
${\mathcal{E}_H^i}=\{h_q^i\}_{q=1}^{N_h^i}$, where each 
relation  tuple
$h_q^i$ contains an arbitrary number of entries and $N_h^i$ is the number of  relation  tuples.

\subsection{Multi-level Knowledge Composition}
{\color{black}As aforementioned, the attribute knowledge refers to the \mbox{attribute-value} pairs of entities mentioned directly in the context, 
whose role is
straightforward
in responding  questions that contain shallow user intentions.}
Meanwhile, the relation knowledge uncovers the correlation between entities, which can benefit the responding to questions that contain complex user intentions.
In light of this, 
we devise the \mbox{multi-level} knowledge composition component 
with the 
\emph{shallow attribute knowledge composition} and \emph{complex relation knowledge composition}.
For simplicity, we temporally omit the script $i$ that indexes the training samples.

\subsubsection{Shallow Attribute Knowledge Composition.}
In this module, we first extract the embeddings  of  the attribute knowledge and multimodal context, respectively.
As for the attribute knowledge, we merge $\mathcal{K}_t^A$ and $\mathcal{K}_v^A$ as a whole $\mathcal{K}_A=[\mathcal{K}_t^A, \mathcal{K}_v^A]$. Thereafter, we treat the set of attribute-value pairs in $\mathcal{K}_A$ as a sequence of tokens, 
and feed it into the position-wise embedding layer of BART to obtain the attribute knowledge embedding $\mathbf{E}_k\in{\mathbb{R}^{N_K \times D}}$, where  $N_K$ is the number of tokens in $\mathcal{K}_A$. 
In the same manner, we can obtain the embedding for the given textual context $\mathcal{T}=[t_1, t_2, \cdots, t_{N_T}]$, denoted as $\mathbf{E}_t\in{\mathbb{R}^{N_T \times D}}$.
For the given visual context, \emph{i.e.,} the
set of images $\mathcal{V}=\{v_1, v_2, \cdots, v_{N_V}\}$, we first utilize ViT-B/32~\cite{DBLP:conf/iclr/DosovitskiyB0WZ21}~(\emph{i.e.,} $\mathcal{B}_v$) pretrained by CLIP~\cite{DBLP:conf/icml/RadfordKHRGASAM21} to encode each image $v_j$, and then employ a fully connected layer and the layer normalization to get each image embedding $\bar{\mathbf{v}}_j$ as follows,
\begin{equation}
    \begin{split}
    \begin{cases}
    \mathbf{v}_j = \mathcal{B}_v({v_j}), j=1,2,\cdots,N_V,\\
    \bar{\mathbf{v}}_j = LN({\mathbf{v}_j^\top}{\mathbf{W}_v^e}+{\mathbf{b}_v^e}),
    \end{cases}
    \end{split}
    \label{eq5}
\end{equation}
where $\mathbf{v}_j$ is the visual representation, extracted by ViT-B/32, of the image $v_j$, and $LN(\cdot)$ refers to the layer normalization. ${\mathbf{W}_v^e}$ and ${\mathbf{b}_v^e}$ are the \mbox{to-be-learned} weight matrix and bias vector of the fully
connected layer, respectively. Finally, let $\mathbf{E}_v=[\bar{\mathbf{v}}_1; \bar{\mathbf{v}}_2; \cdots; \bar{\mathbf{v}}_{N_V}]^\top\in\mathbb{R}^{{N_V}\times{D}}$ denote the  embedding of the visual context.

Thereafter, to derive the attribute knowledge composed response representation, we feed the embeddings of the attribute knowledge and the multimodal context 
into the  encoder $\mathcal{B}_e$ of BART as follows,
\begin{equation}
    \mathbf{T}_t={\mathcal{B}_e}([\mathbf{E}_k, \mathbf{E}_t, \mathbf{E}_v]),
     \label{eq6}
 \end{equation}
where $\mathbf{T}_t\in{\mathbb{R}^{N_b \times D}}$ is the attribute knowledge composed response representation.
$N_b = ({N_K}+{N_T}+{N_V})$ is the total number of tokens.

\subsubsection{Complex Relation Knowledge Composition.}
In this module, we further integrate the complex relation knowledge to 
 refine the  above composed response representation.
To be specific,
we regard each related relation tuple~(\emph{i.e.,} $h_q^i$) as a sequence of words and  feed it into the position-wise embedding layer of BART to
get its embedding $\mathbf{E}_q^h\in {\mathbb{R}^{N_q^e \times D}}$,
where $N_q^e$ is the number of tokens in $h_q^i$.
Thereafter,  we employ the encoder $\mathcal{B}_e$ of BART
to derive the relation tuple representation as follows,
\begin{equation}
    \mathbf{T}_q^h={\mathcal{B}_e}(\mathbf{E}_q^h),
     \label{eq7}
\end{equation}
where $\mathbf{T}_q^h \in {\mathbb{R}^{N_q^e \times D}}$ is the representation of the  relation tuple $h_q^i$.

In fact, different relation  tuples  tend to play 
different  roles in enhancing the textual response generation.
For example, as shown in Figure~\ref{MMD_example}, the relation  tuple  
(Inaniwa Yosuke $\xrightarrow{near}$  Wisma Atria $\xrightarrow{domain}$  mall) conveys more vital clues in generating the response $u_8$ than (Inaniwa Yosuke $\xrightarrow{near}$  Wisma Atria $\xrightarrow{credit cards}$ yes). 
{\color{black}Therefore, 
we resort to the cross-attention mechanism~\cite{DBLP:conf/nips/VaswaniSPUJGKP17} to emphasize relation tuples that are highly related to the composed response representation and  obtain the reorganized relation tuples representation for the given attribute composed representation, 
due to its superior performance in capturing the interaction relation between two items~\cite{wei2019neural,Liu2022DRML,Liu2021IMPGCN,wang2022causal,clvcnet}.}
Specifically, we treat the composed response 
representation 
$\mathbf{T}_t$ obtained in Eqn.(${\ref{eq6}}$) as the  query, and  the relation tuple representations $\mathbf{T}_q^h$ 
as the key and value
as follows,
 \begin{equation}
    \begin{split}
    \begin{cases}
    \mathbf{Q}_m = {\mathbf{T}_t}{\mathbf{W}_Q^m}, \mathbf{K}_m = {\mathbf{T}_h}{\mathbf{W}_K^m},  \mathbf{V}_m = {\mathbf{T}_h}{\mathbf{W}_V^m},\\
    {\bar{\mathbf{T}}_h} = softmax({\mathbf{Q}_m}{\mathbf{K}_m^\top}){\mathbf{V}_m},
    \end{cases}
    \end{split}
    \label{eq8_5} 
\end{equation}
where $\mathbf{T}_h=[\mathbf{t}_h^1, \mathbf{t}_h^2, \cdots, \mathbf{t}_h^{N_h}] \in {\mathbb{R}^{{N_h}\times{D}}}$ denotes the initial representation of all the related relation tuples, and $\mathbf{t}_h^q=avg(\mathbf{T}_q^h)$ refers to the average pooling over the $q$-th relation  tuple  representation ${{\mathbf{T}}^h_q}$ obtained in Eqn.(${\ref{eq7}}$). 
$\mathbf{W}_Q^m$, ${\mathbf{W}_K^m}$, and ${\mathbf{W}_V^m}$ are weight matrices.  $\mathbf{Q}_m\in \mathbb{R}^{N_b \times D}$,   $\mathbf{K}_m \in \mathbb{R}^{N_h \times D}$ and $\mathbf{V}_m \in \mathbb{R}^{N_h \times D}$ are the query, key, value matrices, respectively.
$softmax(\cdot)$ represents the softmax activation function,
and  
{\color{black}${\bar{\mathbf{T}}_h}\in {\mathbb{R}^{{N_b}\times{D}}}$ stands for the reorganized relation tuples representation for the given attribute composed representation.}

Next, to fulfill the relation knowledge composition, instead of merely using the common  residual operation~\cite{DBLP:conf/sigir/Li0YSCZS21}, similar to~\cite{DBLP:conf/sigir/DuY00022a}, 
we adopt the attention mechanism 
to  adaptively fuse the attribute knowledge composed response representation and the  reorganized relation tuples representation. The reason behind is that they may contribute differently towards the ground truth response generation.
According to the attention mechanism, we have,
 \begin{equation}
    \begin{split}
    \begin{cases}
    \mathbf{H}_t = tanh({\mathbf{T}_t}{\mathbf{W}_t}+\mathbf{B}_t),\\
    \mathbf{H}_h = tanh({\bar{\mathbf{T}}_h}{\mathbf{W}_h}+\mathbf{B}_h),\\
    [\mathbf{r}_t, \mathbf{r}_h] = softmax( [\mathbf{H}_t,\mathbf{H}_h]\mathbf{a}),
    \end{cases}
    \end{split}
    \label{eq8}
\end{equation}
where ${\mathbf{W}_t}$ and ${\mathbf{W}_h}$ are weight matrices, while $\mathbf{B}_t$ and $\mathbf{B}_h$ are bias matrices.  
$\mathbf{r}_t\in\mathbb{R}^{N_b}$ and $\mathbf{r}_h\in\mathbb{R}^{N_b}$ denote the  normalized confidence vectors for  the attribute knowledge composed response representation and the relation knowledge representation.
Here, the to-be-learned vector $\mathbf{a}\in \mathbb{R}^{D}$ can be interpreted as the query ``\emph{which part contributes more to the response generation}''.
Ultimately, we reach the 
final multi-level knowledge  composed response 
representation $\mathbf{T}_c \in \mathbb{R}^{{N_b}\times D}$ as follows,
\begin{equation}
    \mathbf{T}_c={\mathbf{r}_t}\odot{\mathbf{T}_t}+{\mathbf{r}_h}\odot{\bar{\mathbf{T}}_h},
     \label{eq9}
\end{equation}
where $\odot$ represents the element-wise multiplication operation.

\subsection{Representation-regularized Response Generation}
Towards the final response generation, one straightforward and commonly used solution 
is to feed the composed  response representation into the BART decoder, and utilize the \mbox{cross-entropy} loss to perform the output-level supervision.
Although the  method is feasible, it only considers the output-level supervision, but 
neglects the potential representation-level regularization. In fact, we can guide the composed response representation learning with the ground truth response representation.
Therefore, 
we devise the
representation-regularized response generation
component with two key modules: 
\emph{representation-level semantic regularization} and \emph{semantic-enhanced response generation}. The former aims to promote the composed response representation learning with a semantic  regularization between the composed response representation and the ground truth response representation. The latter targets at generating the response with not only the original \mbox{multi-level} knowledge composed response representation but also the regularized composed response semantic representation.

\subsubsection{Representation-level Semantic Regularization.}
To conduct the representation-level semantic regularization, 
we first introduce a set of \mbox{to-be-learned} latent variable vectors as queries to 
interact with the multi-level knowledge composed 
 response representation
and the ground truth response representation, respectively, with the goal of projecting them into the same semantic space and deriving their semantic representations. 
Let $\mathbf{P}_g=\{{\mathbf{p}_g^1}, {\mathbf{p}_g^2}, \cdots, {\mathbf{p}_g^{N_P}}\} \in {\mathbb{R}^{N_P\times D}}$ denote the latent variable matrix with $N_P$ variable vectors.
Then for deriving the  multi-level knowledge composed  response semantic representation, we first employ the cross-attention mechanism to distinguish informative representation dimensions, where $\mathbf{P}_g$  is regarded as the query, while the  composed response  representation $\mathbf{T}_c$ is treated as the  key and value. Subsequently, we utilize the multi-layer perceptron~(MLP)~\cite{DBLP:conf/fgr/ZhangLSA98} 
and the residual operation to further enhance the representation generalization and  get the final composed response semantic representation as follows,
 \begin{equation}
    \begin{split}
    \begin{cases}
    \mathbf{Q}_c = {\mathbf{P}_g}{\mathbf{W}_Q^c}, \mathbf{K}_c = {\mathbf{T}_c}{\mathbf{W}_K^c},  \mathbf{V}_c = {\mathbf{T}_c}{\mathbf{W}_V^c},\\
    {\bar{\mathbf{T}}_c} = softmax({\mathbf{Q}_c}{\mathbf{K}_c^\top}){\mathbf{V}_c},\\
    {\tilde{\mathbf{T}}_c} = {\bar{\mathbf{T}}_c}+ f({\bar{\mathbf{T}}_c}),
    \end{cases}
    \end{split}
    \label{eq13}
\end{equation}
where the query $\mathbf{Q}_c$ is projected from the $\mathbf{P}_g$, while the key $\mathbf{K}_c$ and the value $\mathbf{V}_c$ are  projected from the multi-level knowledge composed response representation. ${\bar{\mathbf{T}}_c}$ is  the intermediate  composed response
semantic representation. 
$\mathbf{W}_Q^c$, $\mathbf{W}_K^c$, and  ${\mathbf{W}_V^c}$ are \mbox{to-be-learned} weight matrices.
$f(\cdot)$ refers to the MLP network.
${\tilde{\mathbf{T}}_c}$ is the final 
composed response  semantic representation.

As for the ground truth response,
we first obtain its embedding matrix $\mathbf{E}_r\in \mathbb{R}^{{N_R}\times D}$ by the position-wise embedding layer of BART, where $N_R$ is the total number of its tokens. We then extract its representation with the BART encoder $\mathcal{B}_e$ as follows, 
\begin{equation}
    \mathbf{T}_r={\mathcal{B}_e}(\mathbf{E}_r),
     \label{eq11}
 \end{equation}
where ${\mathbf{T}_r}\in \mathbb{R}^{{N_R} \times D}$ stands for the representation of the ground truth response.
Thereafter, similar to the composed response semantic representation extraction in Eqn.(\ref{eq13}), we resort to the cross-attention mechanism, where we treat $\mathbf{P}_g$ as the  query, and the ground truth response representation $\mathbf{T}_r$  as both key and value. Let ${\tilde{\mathbf{T}}_r}$ be the obtained semantic representation of the ground truth response via the cross-attention mechanism.

Subsequently, to promote the composed response representation learning towards the response generation,
we adopt the Frobenius norm to regularize the composed response semantic representation 
and the ground truth response semantic representation  to be as similar as possible 
as follows,
\begin{equation}
    \mathcal{L}_r={||{\tilde{\mathbf{T}}_r}-{\tilde{\mathbf{T}}_c}||}_F^2,
     \label{eq14}
 \end{equation}
where $||\cdot||_F^2$ is the Frobenius norm.

\subsubsection{Semantic-enhanced Response Generation.}
{\color{black} 
Considering that the user may particularly pay more attention to the entity's attributes to obtain the desired response~\cite{DBLP:conf/sigir/CuiWSHXN19},
we adopt the revised BART decoder~(\emph{i.e.,} $\bar{\mathcal{B}}_d$) introduced by~\cite{DBLP:journals/corr/abs-2207-07934}
 as our decoder, which can distinguish the informative tokens of the related attribute knowledge and adaptively utilize the knowledge to promote the textual response generation.
Compared with the standard BART decoder, the revised BART decoder additionally introduces a \mbox{dot-product} knowledge-decoder  sub-layer between the masked \mbox{multi-head} self-attention mechanism sub-layer and \mbox{multi-head} \mbox{encoder-decoder} attention mechanism sub-layer.}
To be specific, we feed the  knowledge composed response representation  $\mathbf{T}_c$ and the attribute knowledge embedding $\mathbf{E}_k$ into $\bar{\mathcal{B}}_d$ as follows,
\begin{equation}
    \bar{\mathbf{z}}_j^{dec} = \bar{{\mathcal{B}_d}}(\mathbf{T}_c, \mathbf{E}_k, {\tilde{y}_1}, {\tilde{y}_2}, \cdots, {\tilde{y}_{j-1}}),
    \label{eq_revie_decoder}
\end{equation} 
where $\bar{\mathbf{z}}_j^{dec}$ is the latent representation for generating the $j$-th token learned by the revised BART decoder $\bar{\mathcal{B}}_d$.

{\color{black}
Thereafter, different from previous studies that directly predict the response token distribution based on $\bar{\mathbf{z}}_j^{dec}$, 
we further incorporate the regularized composed response semantic representation to  promote the  textual response generation.} 
 Considering that different semantic dimensions can contribute differently towards the response generation,  we also employ the \mbox{cross-attention} mechanism to obtain 
 the reorganized composed response semantic representation for
 $\bar{\mathbf{z}}_j^{dec}$. Specifically, we treat the latent representation $\bar{\mathbf{z}}_j^{dec}$ as the query,  and the composed response
semantic representation  ${\tilde{\mathbf{T}}_c}$ as the key and value as follows,
 \begin{equation}
    \begin{cases}
    \mathbf{q}_d = ({\bar{\mathbf{z}}_j^{dec}})^{\top}{\mathbf{W}_Q^d}, \mathbf{K}_d = {\tilde{\mathbf{T}}_c}{\mathbf{W}_K^d},  \mathbf{V}_d = {\tilde{\mathbf{T}}_c}{\mathbf{W}_V^d},\\
    {\hat{\mathbf{t}}_c} = softmax({\mathbf{q}_d^\top}{\mathbf{K}_d^\top}){\mathbf{V}_d},
    \end{cases}
    \label{eq15}
\end{equation}
where $\mathbf{q}_d$ is the query vector, while $\mathbf{K}_d$ and $\mathbf{V}_d$ are the  key and value matrices, respectively.
$\mathbf{W}_Q^d$, $\mathbf{W}_K^d$, and ${\mathbf{W}_V^d}$ are weight matrices.
${\hat{\mathbf{t}}_c}$ is the refined  composed response semantic representation. 
Thereafter, we can obtain the semantic-enhanced latent representation by fusing $\bar{\mathbf{z}}_j^{dec}$ and ${\hat{\mathbf{t}}_c}$ as follows,
\begin{equation}
    {{\hat{\mathbf{z}}_j^{dec}}} = LN(\bar{\mathbf{z}}_j^{dec}+\hat{\mathbf{t}}_c),
\end{equation} 
where ${{\hat{\mathbf{z}}_j^{dec}}}$ is the semantic-enhanced latent representation for generating the $j$-th response token. Specifically,  
we can derive the $j$-th token probability distribution based on ${\hat{\mathbf{z}}_j^{dec}}$ as follows,
\begin{equation}
    \hat{{\mathbf{y}}}_j = softmax(({\hat{\mathbf{z}}_j^{dec}})^{\top} {\mathbf{W}_y}+\mathbf{b}_y),
     \label{eq16}
 \end{equation}
 where ${\mathbf{W}_y}$ and $\mathbf{b}_y$ denote the weight matrix and bias vector, respectively. $\hat{{\mathbf{y}}}_j$  represents the predicted probability distribution for the $j$-th token of the response.
Notably, to avoid error accumulation, in practice, we utilize  the 
semantic representation of
ground truth response
${\tilde{\mathbf{T}}_r}$ instead of the  composed response semantic representation 
${\tilde{\mathbf{T}}_c}$ during the training phase.

Ultimately, 
following existing methods~\cite{10.1145/3343031.3350923,DBLP:conf/mm/LiaoM0HC18}, we  use the cross entropy loss~\cite{DBLP:journals/pr/LiL93}  to fulfill the output-level supervision  
as follows,
\begin{equation}
    \mathcal{L}_{CE} = -{\frac{1}{N_R}}{{\sum}_{n=1}^{N_R}{log({\hat{\mathbf{y}}}_n[t*])}},
     \label{eq17}
\end{equation}
where ${\tilde{\mathbf{y}}}_n[t*]$ denotes the element of ${\tilde{\mathbf{y}}}$ that corresponds to the $n$-th token of the ground truth response $\mathcal{R}$, and $N_R$ is the number of tokens in $\mathcal{R}$. Notably, the loss is defined for a single sample.
In the end, we can reach the final objective  function for the textual response generation as follows,
\begin{equation}
    \mathcal{L} = \lambda{\mathcal{L}_{CE}} + \gamma{\mathcal{L}_r} + \beta{||\boldsymbol{\Theta}_F||^2_F},
     \label{eq18}
\end{equation}
where $\lambda$, $\gamma$, and $\beta$ are the non-negative hyper-parameters.   $\boldsymbol{\Theta}_F$ denotes the set of parameters of the proposed \mbox{MDS-S$^2$}.
\section{Experiment}

\subsection{Experiment Setting}
\textbf{Dataset.} Previous studies mainly evaluate their models on the public  dataset MMD constructed by Saha et al.~\cite{DBLP:conf/aaai/SahaKS18} from the fashion domain.
However, similar to~\cite{DBLP:journals/corr/abs-2207-07934}, we did not evaluate our model on this dataset owing to 
the fact that MMD only allows the 
attribute knowledge acquisition.
In contrast, we utilized the public dataset MMConv~\cite{10.1145/3404835.3462970}, which supports the 
knowledge acquisition from both the attribute and relation perspectives.
The MMConv dataset consists of $5,106$  conversations between users and agents covering five domains: \textit{Food}, \textit{Hotel}, \textit{Nightlife}, \textit{Shopping mall}, and \textit{Sightseeing}.
Thereinto, the number of single-modality and \mbox{multi-modality} dialogues are $751$ and $4,355$, respectively, where the average number of turns are $7.1$ and $7.9$.
Besides, the knowledge base contains $1,771$ knowledge entities, where each entity is equipped with 
a set of attribute-value pairs 
and a few images.
The average number of attribute-value pairs and images are $13.7$ and $64.3$, respectively.

\textbf{Implementation Details.} Following the original setting in MMConv, we divided dialogues into three chunks: $3,500$ for training, $606$ for validation, and $1,000$ for testing.
Similar to existing studies~\cite{DBLP:journals/corr/abs-2207-07934,DBLP:conf/sigir/CuiWSHXN19}, we regarded each utterance of agents as a ground truth  response and employed its former two-turn utterances as the given context.
{\color{black}We adopted the pretrained BART-large\footnote{https://huggingface.co/facebook/bart-large.}~\cite{DBLP:conf/emnlp/WolfDSCDMCRLFDS20} model with $12$ layers for the encoder and decoder, respectively.}
As for optimization, we utilized the adaptive moment estimation optimizer and settled the learning rate as \mbox{$1e$-$5$}.
In addition, we fine-tuned the proposed \mbox{MDS-S$^2$}  based on the training and validation dataset with $100$ epochs, and reported the performance on the testing dataset.
Besides, we implemented our model by  Pytorch~\cite{10.5555/3454287.3455008} and deployed all experiments on a server equipped with $8$ NVIDIA $3090$ GPUs.
{\color{black}{Following existing studies~\cite{DBLP:journals/corr/abs-2207-07934,10.1145/3343031.3350923}, we utilized BLEU-$N$~\cite{papineni-etal-2002-bleu} where $N$ ranges from $1$ to $4$, and Nist~\cite{10.5555/1289189.1289273} as evaluation metrics.
}}

\subsection{Model Comparison (RQ1)}

To verify the effectiveness of our proposed \mbox{MDS-S$^2$}, we chose the following state-of-the-art models on multimodal task-oriented dialog systems as baselines.

\begin{table}[!t]
    \centering
    \caption{Performance comparison among different methods in terms of BLEU-\emph{N}~($\%$) and Nist. ``Improve.$\uparrow$'': the relative improvement by our model over the best baseline. {\color{black}The best results are in boldface, and the second best are underlined.}}
    \vspace{-0.5em}
    \begin{tabular}{|l||r|r|r|r|r|}
    \hline
        Models & BLEU-1 & BLEU-2 & BLEU-3 & BLEU-4 & Nist \\ \hline
        MHRED & $15.02$ & $6.66$ & $4.24$ & $2.94$ & $0.9529$ \\ \hline
        KHRED & $18.29$ & $8.28$ & $4.98$ & $3.36$ & $1.1189$ \\ \hline
        LARCH & $20.86$ & $11.33$ & $7.58$ & $5.58$ & $1.3400$ \\ \hline
        MATE & $30.45$ & $22.06$ & $17.05$ & $13.41$ & $2.3426$ \\ \hline
        UMD & $31.14$ & $21.87$ & $17.12$ & $13.82$ & $2.5290$ \\ \hline
        TREASURE & $34.75$ & $24.82$ & $18.67$ & $14.53$ & $2.4398$ \\ \hline
        {DKMD} & \underline{${39.59}$} & \underline{$31.95$} & \underline{${27.26}$} & \underline{${23.72}$} & \underline{${4.0004}$} \\ \hline
        \textbf{MDS-S$^2$} & $\textbf{41.40}$ & $\textbf{32.91}$ & $\textbf{27.74}$ & $\textbf{23.89}$ & $\textbf{4.2142}$ \\ \hline
         Improve.$\uparrow$ & $4.57\%$ & $3.00\%$ & $1.76\%$ & $0.72\%$ & $5.34\%$ \\ \hline
    \end{tabular}
    \vspace{-1.5em}
    \label{rq1}
\end{table}

\begin{figure*}[!t]
    \centering
    \vspace{0em}
     \subfigure[Case $1$.]{
      \includegraphics[scale=0.55
      ]{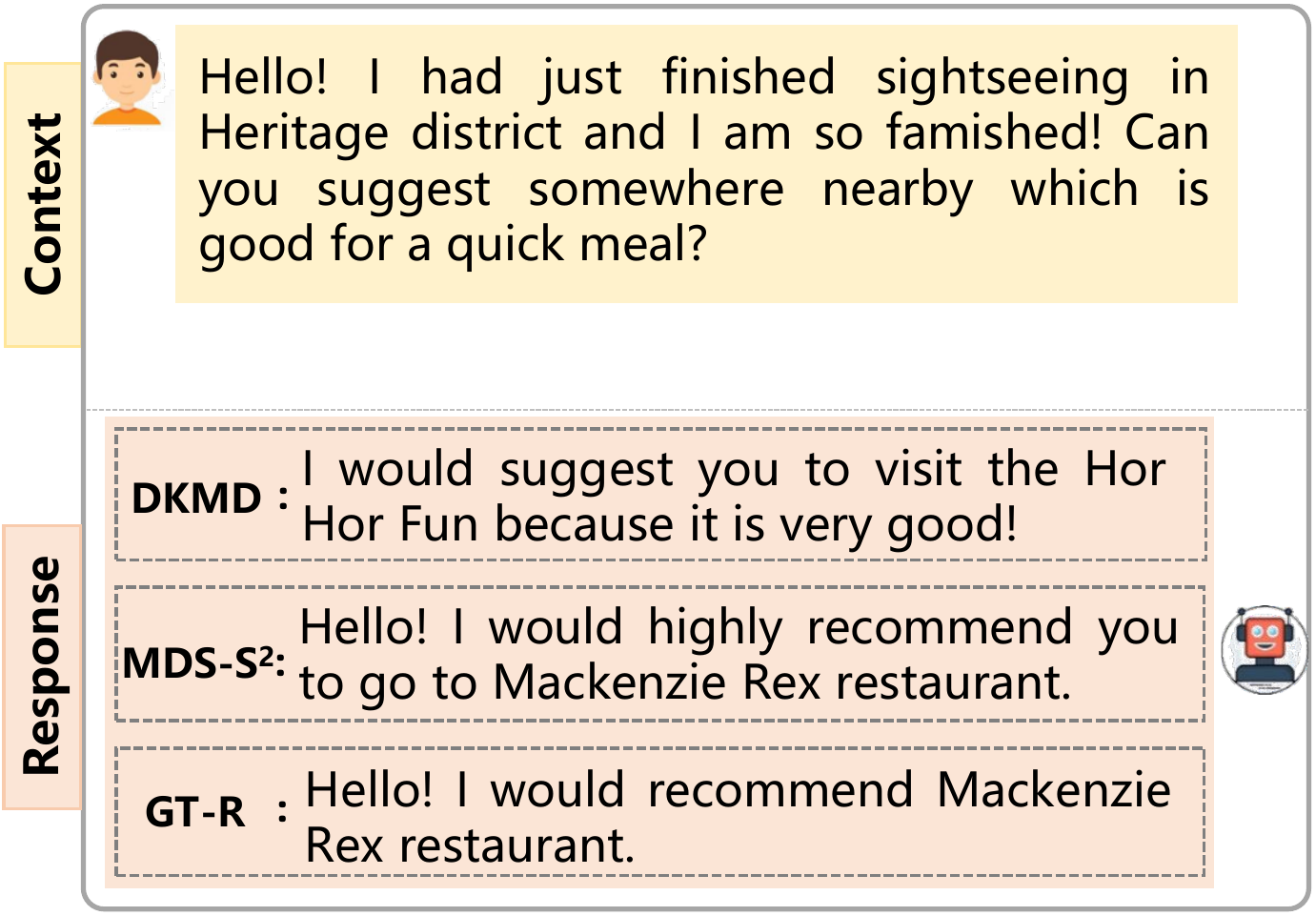}
       \label{rq1:subfig1}
       }
     \subfigure[Case $2$.]{
       \includegraphics[scale=0.55]{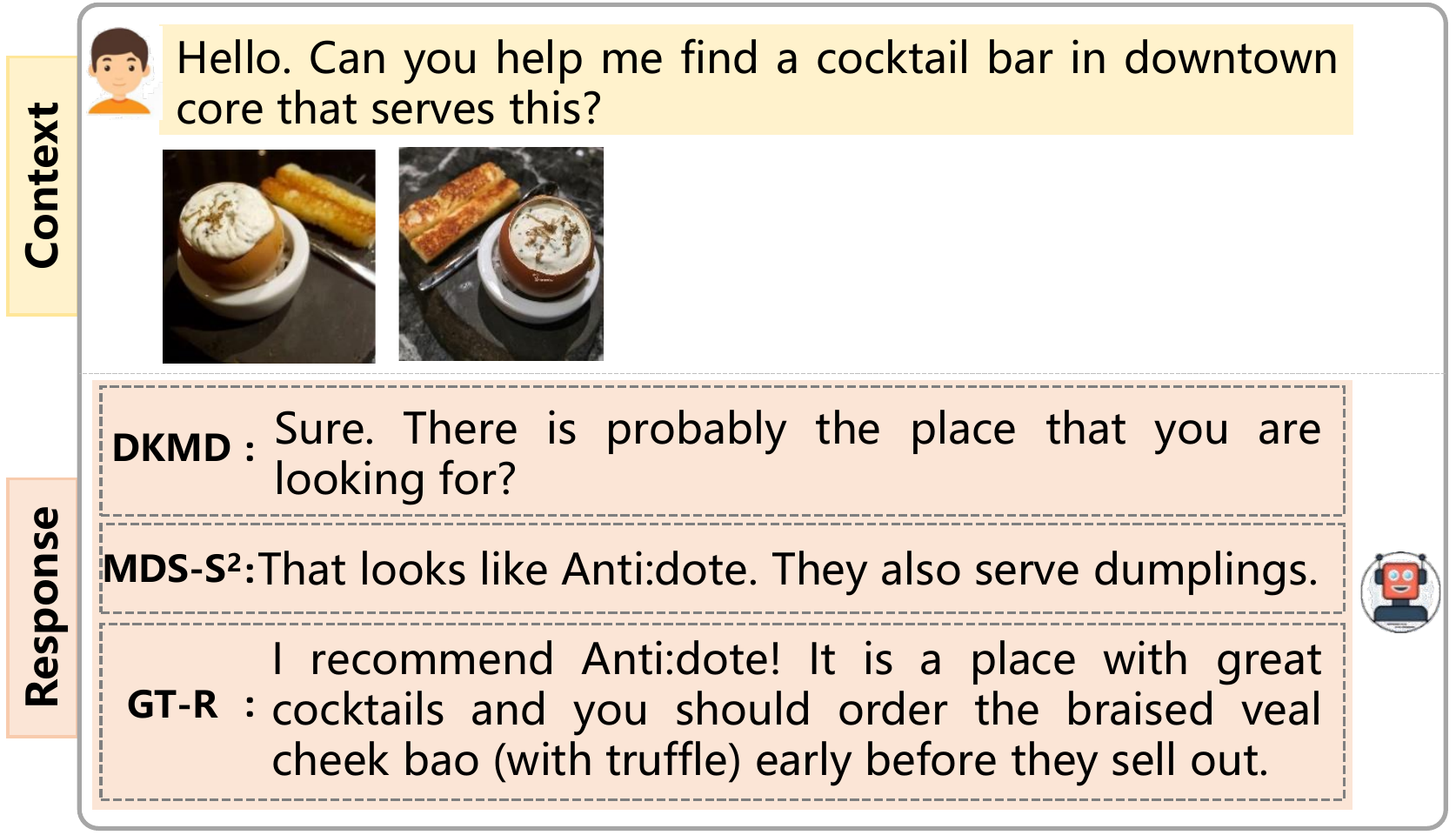}
       \label{rq1:subfig2}
       }
    \vspace{-0.8em}
    \caption{{\color{black}Comparison between our \mbox{MDS-S$^2$} and DKMD on  two testing dialog pairs. ``GT-R'' refers to the ground truth response.}}
    \vspace{-0.5em}
    \label{Rq1_baseline_case}
\end{figure*}

\begin{itemize}[leftmargin=*]
    \item \textbf{MHRED}~\cite{DBLP:conf/aaai/SahaKS18} is the first study on the multimodal task-oriented dialog systems with a hierarchical encoder and a decoder. Thereinto, the hierarchical encoder consists of two levels of the gated recurrent units (GRU)~\cite{DBLP:journals/corr/ChungGCB14},   modeling the utterance and context, respectively. This  baseline does not consider the knowledge  as well as the \mbox{representation-level} regularization.
    \item \textbf{KHRED}~\cite{DBLP:journals/corr/abs-2207-07934} is extended from MHRED by integrating the attribute knowledge. Specifically,  this method employs the memory network to encode the attribute knowledge, and then uses the GRU-based decoder to generate the textual response.
    \item \textbf{LARCH}~\cite{DBLP:journals/tip/NieJWWT21} employs a multimodal hierarchical graph to encode the given dialog context, where each word, image, sentence, utterance, dialog pair, and the session is regarded as a node. In addition, considering the pivotal role of knowledge in multimodal  dialog systems, this method
    integrates the attribute knowledge with a memory network.
    \item \textbf{MATE}~\cite{DBLP:conf/mm/LiaoM0HC18} first resorts to the Transformer network to explore the semantic relation between the textual context and the visual context, and thus  enhances the context representation. Thereafter, the method  utilizes the Transformer-based decoder to generate the textual response.
    \item \textbf{UMD}~\cite{DBLP:conf/sigir/CuiWSHXN19} 
    utilizes a hierarchy-aware tree encoder to capture the taxonomy-guided attribute-level visual representation, and a multimodal factorized bilinear pooling layer to obtain the utterance representation.  
    \item \textbf{TREASURE}~\cite{DBLP:conf/mm/ZhangLGLWN21} presents an attribute-enhanced textual encoder, which integrates the attribute knowledge into the utterance representation. Besides, the method adopts a graph attention network to capture the semantic relation among utterances and obtain the context representation.
    \item    \textbf{DKMD}~\cite{DBLP:journals/corr/abs-2207-07934} presents a  dual knowledge-enhanced generative pretrained language model, where BART is adopted as the backbone. In particular, the method only explores  the textual  and visual context related attribute knowledge, overlooking the relation knowledge and the representation-level regularization.
\end{itemize}


Table~\ref{rq1} summarizes the performance comparison among different methods with respect to different evaluation metrics.
From this table, we can draw the following observations. 
\mbox{1) Our} proposed \mbox{MDS-S$^2$} consistently surpasses all the baselines, exhibiting the 
superiority
of the proposed network. 
In a sense, this suggests that it is reasonable to integrate the multi-level knowledge composition as well as the representation-level regularization.
2) Our proposed \mbox{MDS-S$^2$} outperforms all the baselines 
that only consider the attribute knowledge   (\emph{i.e.,} DKMD, TREASURE, UMD, LARCH, and KHRED), which indicates the advantage of 
simultaneously 
incorporating both
the attribute knowledge and relation knowledge
in multimodal \mbox{task-oriented} dialog systems.
\mbox{3) MHRED} gets the worst performance compared to other methods. This may be due to the fact that MHRED overlooks both the dual semantic knowledge~(\emph{i.e.,} attribute and relation knowledge)
 and the representation-level  regularization.
And 4) both \mbox{MDS-S$^2$} and DKMD exceed other baselines, 
which confirms the benefit of exploiting the generative language model in the context of multimodal  \mbox{task-oriented} dialog systems.

{\color{black}To intuitively verify the effectiveness of the proposed \mbox{MDS-S$^2$}, we randomly selected two testing  dialog pairs, and  exhibited the responses generated by the \mbox{MDS-S$^2$} and the best baseline DKMD due to the space limitation in Figure~\ref{Rq1_baseline_case}. 
As can be seen, 
our proposed \mbox{MDS-S$^2$} outperforms DKMD in the \emph{case 1} that may involve the relation knowledge.
Meanwhile, 
In case~$2$ that does not need the complicated relation knowledge,
we found that our proposed \mbox{MDS-S$^2$} can generate the appropriate response, while DKMD fails. } 
One possible explanation is that our proposed \mbox{MDS-S$^2$} can promote the composed response representation learning
with the \mbox{representation-level} regularization.

\subsection{On Dual Semantic  Knowledge (RQ2)}
To explore the roles of dual semantic knowledge
in multimodal  dialog systems, we designed the following five derivations.
\mbox{1) \textbf{w/o-rel.}} To illustrate the importance of the relation knowledge, we only kept the shallow attribute knowledge composition and utilized the composed response representation $\mathbf{T}_t$ in Eqn.(${\ref{eq6}}$) as the input for Eqn.(${\ref{eq_revie_decoder}}$).
2)~{\textbf{w/o-att.}} To verify the importance of the attribute knowledge, we removed both the shallow attribute knowledge composition
and the attribute knowledge from Eqn.(${\ref{eq_revie_decoder}}$) by using the standard 
 BART decoder.
3)~{\textbf{w/o-att-com.}} To exhibit the necessity of the shallow attribute knowledge composition, we disabled the attribute knowledge in  Eqn.(${\ref{eq6}}$).
4)~{\textbf{w/o-att-dec.}} To demonstrate the necessity of incorporating the attribute knowledge into the decoder, we replaced the knowledge revised decoder~(\emph{i.e.,} $\bar{{\mathcal{B}_d}}$) with the original decoder of BART in Eqn.(${\ref{eq_revie_decoder}}$).
5)~{\textbf{w/o-dual-k.}} To illustrate the role of knowledge towards the textual response generation, we removed all the attribute and relation knowledge from our proposed \mbox{MDS-S$^2$}.

\begin{table}[!b]
    \centering
    \vspace{-1em}
    \caption{Ablation study results on the dual semantic  knowledge in terms of BLEU-\emph{N}~($\%$) and Nist.}
    \vspace{-1em}
    \begin{tabular}{|l||r|r|r|r|r|}
    \hline
        Models & BLEU-1 & BLEU-2 & BLEU-3 & BLEU-4 & Nist \\ \hline
        w/o-rel & $36.33$ & $27.50$ & $22.60$ & $19.02$ & $3.5398$ \\ 
        \hline
        w/o-att & $33.58$ & $24.80$ & $20.16$ & $16.93$ & $2.9815$ \\ \hline
        w/o-att-com & $34.82$ & $26.28$ & $21.62$ & $18.28$ & $3.3089$ \\  \hline
        w/o-att-dec & $39.54$ & $31.68$ & $26.97$ & $23.46$ & $4.0662$ \\ \hline
        w/o-dual-k & $32.65$ & $24.30$ & $19.85$ & $16.69$ & $2.9856$ \\ 
        \hline
        \textbf{MDS-S$^2$} & $\textbf{41.40}$ & $\textbf{32.91}$ & $\textbf{27.74}$ & $\textbf{23.89}$ & $\textbf{4.2142}$ \\ \hline
    \end{tabular}
    \label{rq2}
\end{table}

Table~\ref{rq2} demonstrates  the performance comparison between our proposed \mbox{MDS-S$^2$} and its above derivations. From this table, we had the following observations.
1) Our proposed \mbox{MDS-S$^2$} outperforms w/o-rel,  w/o-att, and w/o-dual-k. Besides, disabling all  knowledge~(\emph{i.e,} w/o-dual-k) results in the worst performance. 
It exhibits that removing either the attribute or the relation knowledge will hurt the performance of \mbox{MDS-S$^2$} to some extent. 
This reconfirms the superiority of considering the dual semantic knowledge in  multimodal task-oriented dialog systems.
2) Both w/o-att and \mbox{w/o-att-com} perform worse than w/o-rel, suggesting that the attribute knowledge contributes more to textual response generation than the relation knowledge.
{\color{black} This may be due to the fact that in most cases,  users want to learn the attribute information of certain known entities, and the attribute knowledge is enough for responding such cases.
}
And \mbox{3) our} proposed \mbox{MDS-S$^2$} surpasses both w/o-att-com and w/o-att-dec.
The rationale behind is that the attribute knowledge integrated in the shallow  attribute knowledge composition can  facilitate the composed response representation learning, while that in the decoder is able to explicitly incorporate the attribute knowledge in the decoder phrase, both benefiting the final textual response generation.

\begin{figure}[!t]
    \centering
    \includegraphics[scale=0.37]{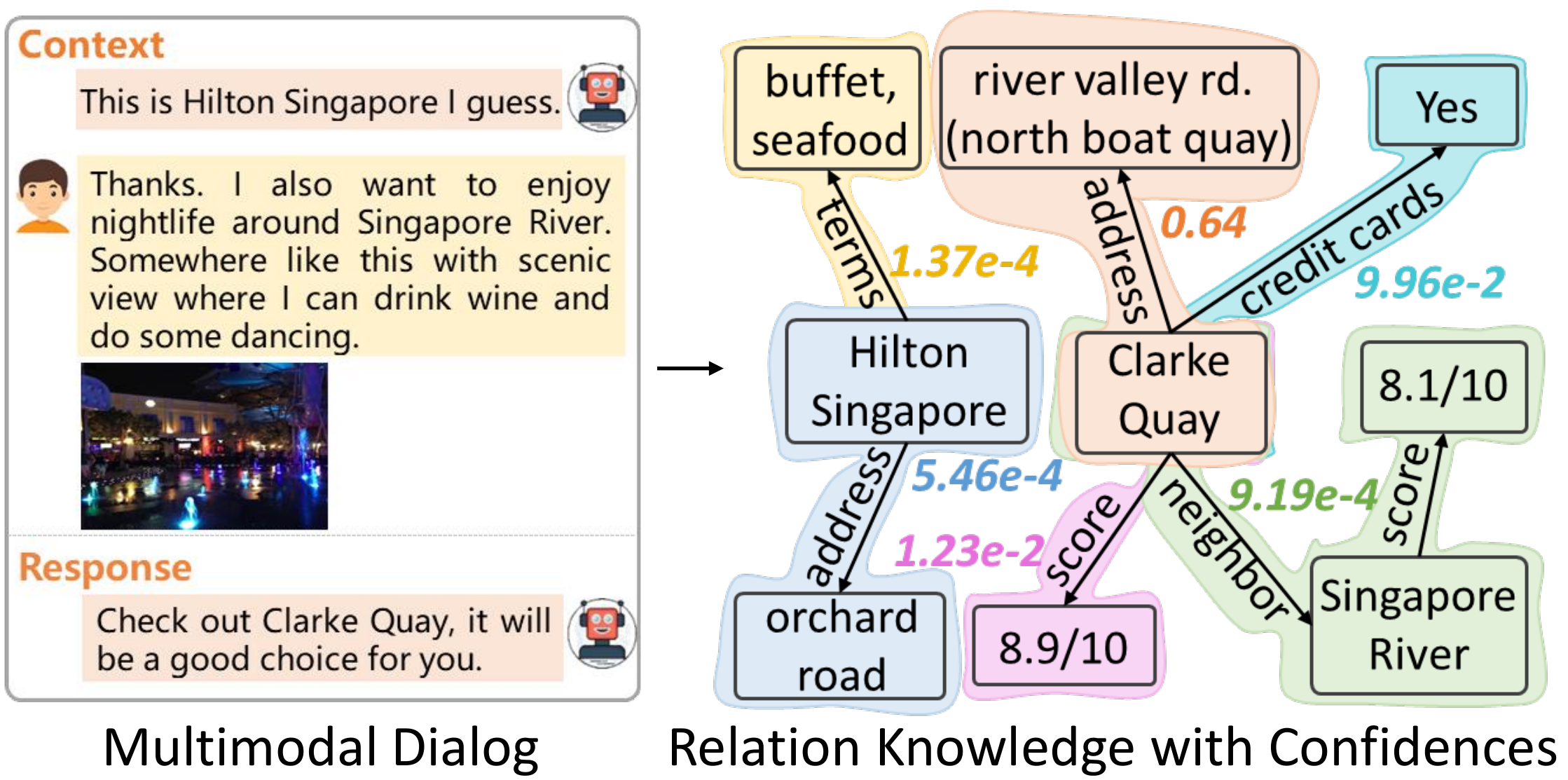}
    \vspace{-1.5em}
    \caption{Illustration of the learned confidences for the relation knowledge.}
    \vspace{-1.5em}
    \label{Rq2_know_case}
\end{figure}

To obtain deeper insights into the relation knowledge,
we performed the case study on the relation knowledge confidence assignment with a testing multimodal dialog in Figure~\ref{Rq2_know_case}. Due to the limited space, we only showed a few essential relation tuples related to the given context.
Different color shadows correspond to different relation  tuples.
{As we can see, towards the textual response generation, \mbox{MDS-S$^2$} assigns the highest confidence to the  relation  tuple ``Clarke Quay $\xrightarrow{address}$ river valley rd. (north boat quay)'', as compared to other tuples.
{\color{black}Checking the given multimodal context, we found that the user provides an image and wants to find a similar place with  a scenic view around Singapore River. In this case, the address of the entity~(\emph{i.e.,} ``river valley rd. (north boat quay)'') indicates that it may be near the river, which is  instructive for the proper response generation. }
In light of this, the confidence assignment of our model regarding relation tuples for the given multimodal context is reasonable. 
}
This suggests that our model can  identify the informative relation  tuples to enhance the textual response generation in multimodal \mbox{task-oriented} dialog systems.

\begin{table}[!b]
    \centering
    \vspace{-1em}
    \caption{Ablation study results on the representation-level regularization  in terms of BLEU-\emph{N}~($\%$) and Nist.}
    \vspace{-1em}
    \begin{tabular}{|l||r|r|r|r|r|}
    \hline
        Models & BLEU-1 & BLEU-2 & BLEU-3 & BLEU-4 & Nist \\ \hline
        w/o-regular & $39.13$ & $31.28$ & $26.54$ & $23.02$ & $3.9602$ \\ \hline
        w/o-sem-dec & $37.63$ & $29.39$ & $24.50$ & $20.91$ & $3.6561$ \\ \hline
        \textbf{MDS-S$^2$} & $\textbf{41.40}$ & $\textbf{32.91}$ & $\textbf{27.74}$ & $\textbf{23.89}$ & $\textbf{4.2142}$ \\ \hline
    \end{tabular}
    \label{rq3}
\end{table} 

\subsection{On Representation-level Regularization (RQ3)}
To thoroughly verify the effectiveness of the \mbox{representation-level} regularization, we designed two  derivatives. 
\mbox{1) \textbf{w/o-regular}.} To verify the importance of the representation-level regularization, we removed it from our model. Consequently, the  composed response semantic representation will not be used.
\mbox{2) \textbf{w/o-sem-dec}.} To exhibit the benefit of injecting the multi-level
knowledge composed response semantic representation into the BART decoder, we kept the representation-level regularization but disabled the composed response semantic representation integration in the decoder.

Table~\ref{rq3} summarizes the performance of our \mbox{MDS-S$^2$} and its derivatives.
As can be seen, our proposed \mbox{MDS-S$^2$} outperforms w/o-regular, which suggests the necessity of conducting the representation-level
regularization
in the context of multimodal task-oriented dialog systems. 
Besides, we found that w/o-sem-dec performs worse than our proposed \mbox{MDS-S$^2$}. 
This may be due to that the composed response semantic representation  integrated into the decoder phrase  can directly promote the textual response generation and thus enhance the performance.
Thirdly, w/o-regular underperforms w/o-sem-dec, revealing that it is reasonable to conduct the representation-level regularization, which contributes to 
regularize the composed response semantic representation to be similar to the ground truth response semantic representation.

\begin{figure}[!t]
    \centering
     \subfigure[MDS-S$^2$.]{
      \includegraphics[scale=0.19]{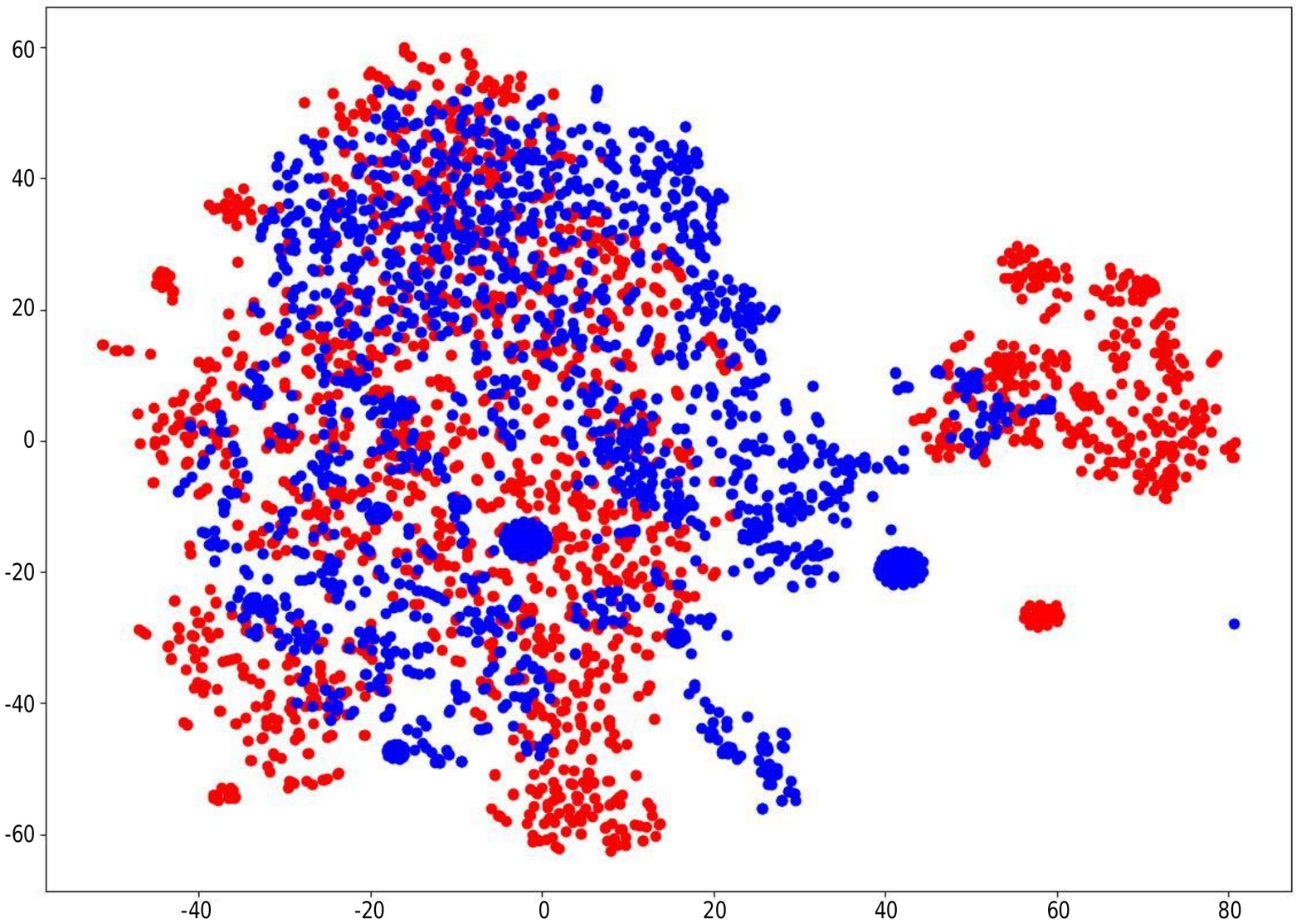}
       \label{variable:subfig1}
       }
     \subfigure[w/o-regular.]{
       \includegraphics[scale=0.19]{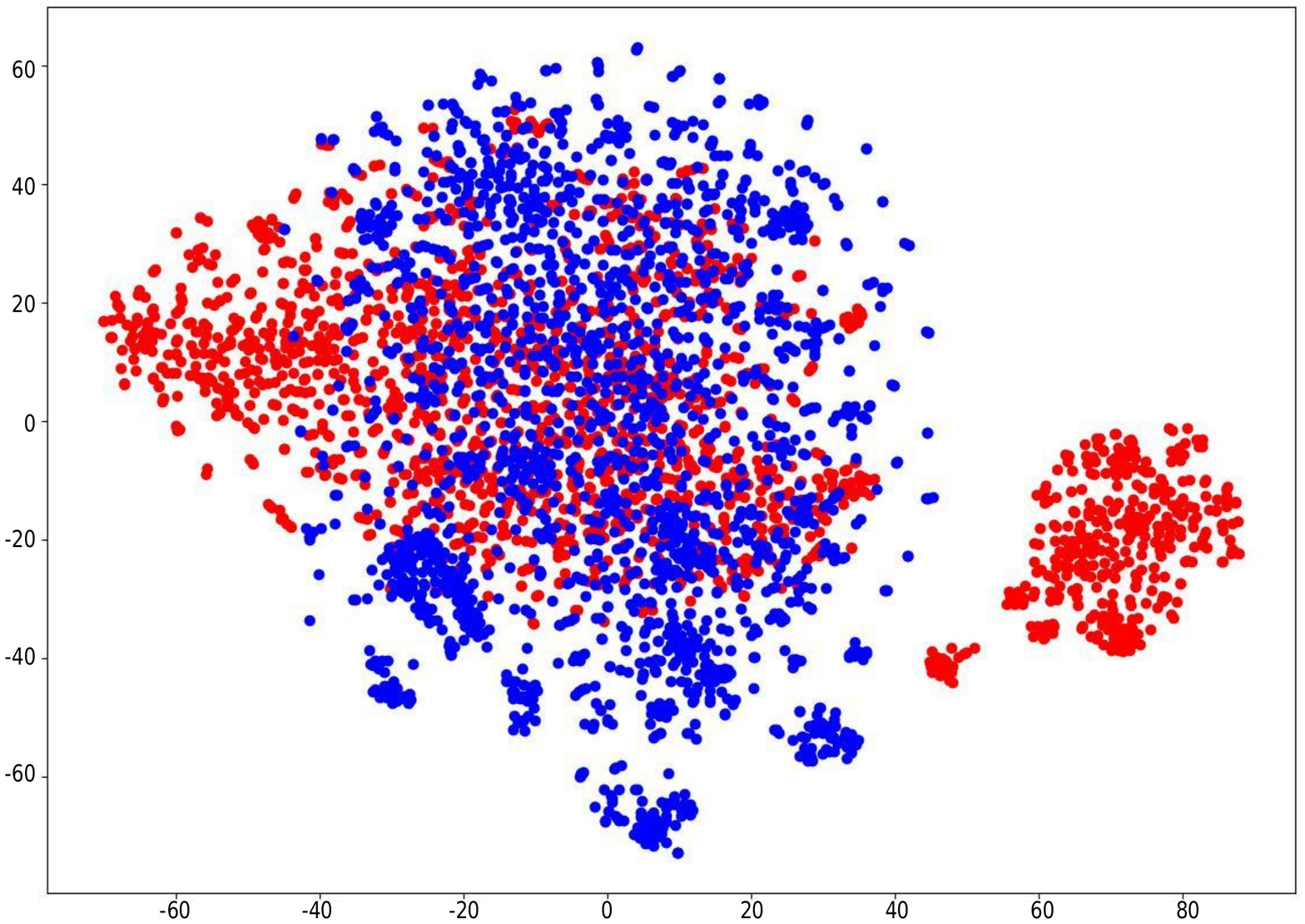}
       \label{att:subfig2} 
       }
    \vspace{-1.5em}
    \caption{Visualization of the  composed response  representation distribution (red points) as well as the ground truth response  representation distribution (blue points) learned by our \mbox{MDS-S$^2$} and its derivative w/o-regular. 
    }
    \vspace{-1.5em}
    \label{rq3_variable}
\end{figure}

To intuitively reflect the effectiveness of the \mbox{representation-level} regularization,  we randomly sampled $2,000$  multimodal dialogs, and visualized  the multi-level knowledge composed response representation as well as the ground truth response representation learned by our \mbox{MDS-S$^2$} and w/o-regular with the help of tSNE~\cite{DBLP:conf/vissym/RauberFT16} in Figure~\ref{rq3_variable}.  
The red points illustrate the  composed response representation, and the blue ones denote  the  ground truth  response representation.
{\color{black}As we can see, 
the distribution of the  multi-level knowledge composed response representation and that of the ground truth response representation achieved by \mbox{MDS-S$^2$} is more consistent as compared to that obtained by  w/o-regular.}
This validates the effect of our proposed 
\mbox{representation-level} regularization in capturing the meaningful information from the composed response representation
towards the textual response generation.
{\color{black}In addition, as for our \mbox{MDS-S$^2$},  we found that there is still a separate region where the two representation distributions are not well aligned.  
Checking these samples, we learned that they tend to involve open questions 
(\emph{e.g.,} ``any tips when visiting there?''), for which the effect of the dual semantic knowledge is limited. 
}

\section{Conclusion and Future Work}
{\color{black}In this work, we investigate the textual response generation task in multimodal task-oriented dialog systems
and propose a novel multimodal dialog system, named \mbox{MDS-S$^2$}.}
Extensive experiments on a public dataset  have validated the effectiveness of the proposed \mbox{MDS-S$^2$}.
{\color{black}Interestingly, we observe that the attribute knowledge and the relation knowledge are both conducive to the textual response generation. 
}
Besides, the representation-level regularization does help in guiding the composed response representation learning with the ground truth response and should be taken into account.
As aforementioned, the public MMConv dataset covers dialogs of multiple domains {\color{black}(\textit{e.g.,} \textit{Food}, \textit{Hotel},  and \textit{Shopping mall})}. Currently, we did not explore the domain information of each dialog towards the textual response generation. In the future, we plan to explore the 
semantic transition among different domain topics in the multimodal context and further enhance the response generation performance of multimodal dialog systems. 

\begin{acks}
{\color{black}This work is supported by the National Key Research and Development Project of New Generation Artificial Intelligence, No.:2018AAA0102502, the Shandong Provincial Natural Science Foundation (No.:ZR2022YQ59),  the National Natural Science Foundation of China, No.:U1936203 and No.:62236003; Shenzhen College Stability Support Plan, No.: GXWD20220817144428005. Xiaolin Chen was supported by the China Scholarship Council for 1 year study at the National University of Singapore.}
\end{acks}

\clearpage
\small

\bibliographystyle{ACM-Reference-Format}
\bibliography{reference}

\end{document}